\documentclass[10pt,twocolumn,letterpaper]{article}

\usepackage{acpr}
\usepackage{times}
\usepackage{epsfig}
\usepackage{graphicx}
\usepackage{amsmath}
\usepackage{amssymb}

\usepackage{tabularx}
\usepackage{tabu}
\usepackage{hhline}

\usepackage{nopageno}
\usepackage{array,colortbl}
\usepackage{makecell}

\newcolumntype{?}{!{\vrule width 1pt}}

\usepackage[pagebackref=true,breaklinks=true,letterpaper=true,colorlinks,bookmarks=false]{hyperref}

\acprfinalcopy % *** Uncomment this line for the final submission

\ifacprfinal\fi
\begin{document}

%%%%%%%%% TITLE
\title{Real-time Sign Language Fingerspelling Recognition using \\Convolutional Neural Networks from Depth map}

\author{Byeongkeun Kang \hspace{1cm} Subarna Tripathi \hspace{1cm} Truong Q. Nguyen \\
Department of Electrical and Computer Engineering \\
University of California, San Diego\\
{\tt\small {bkkang, stripathi, tqn001}@ucsd.edu}
}

\maketitle
%\thispagestyle{empty}

%%%%%%%%% ABSTRACT
\begin{abstract}
Sign language recognition is important for natural and convenient communication between deaf community and hearing majority. We take the highly efficient initial step of automatic fingerspelling recognition system using convolutional neural networks (CNNs) from depth maps. In this work, we consider relatively larger number of classes compared with the previous literature. We train CNNs for the classification of 31 alphabets and numbers using a subset of collected depth data from multiple subjects. While using different learning configurations, such as hyper-parameter selection with and without validation, we achieve 99.99\% accuracy for observed signers and 83.58\% to 85.49\% accuracy for new signers. The result shows that accuracy improves as we include more data from different subjects during training. The processing time is 3 ms for the prediction of a single image. To the best of our knowledge, the system achieves the highest accuracy and speed. The trained model and dataset is available on our repository\footnote{\url{https://github.com/byeongkeun-kang/FingerspellingRecognition}}.
\end{abstract}

%%%%%%%%% BODY TEXT
\section{Introduction}
Sign language recognition is important for natural and convenient communication between deaf community and hearing majority. Currently, most communications between two communities highly rely on human-based translation services. However, this is inconvenient and expensive as human expertise is involved. Therefore, automatic sign language recognition aims to understand the meaning of signs without the assistance from experts. Then it can be translated to sound or text based on end users' needs. We believe that sign language recognition is important for providing equal opportunity to every person and improving public welfare.

Sign language recognition is still a challenging problem despite of many research efforts during the last few decades~\cite{star98, coop11}. It requires the understanding of combination of multi-modal information such as hand pose and movement, facial expression, and human body posture. Also, sign language has at least thousands of words including very similar hand poses while gesture recognition generally includes a small set of well specified gestures. Moreover, even same signs have significantly different appearances for different signers and different viewpoints.

\begin{figure}[t]
\begin{center}
   \includegraphics[width=1.0\linewidth]{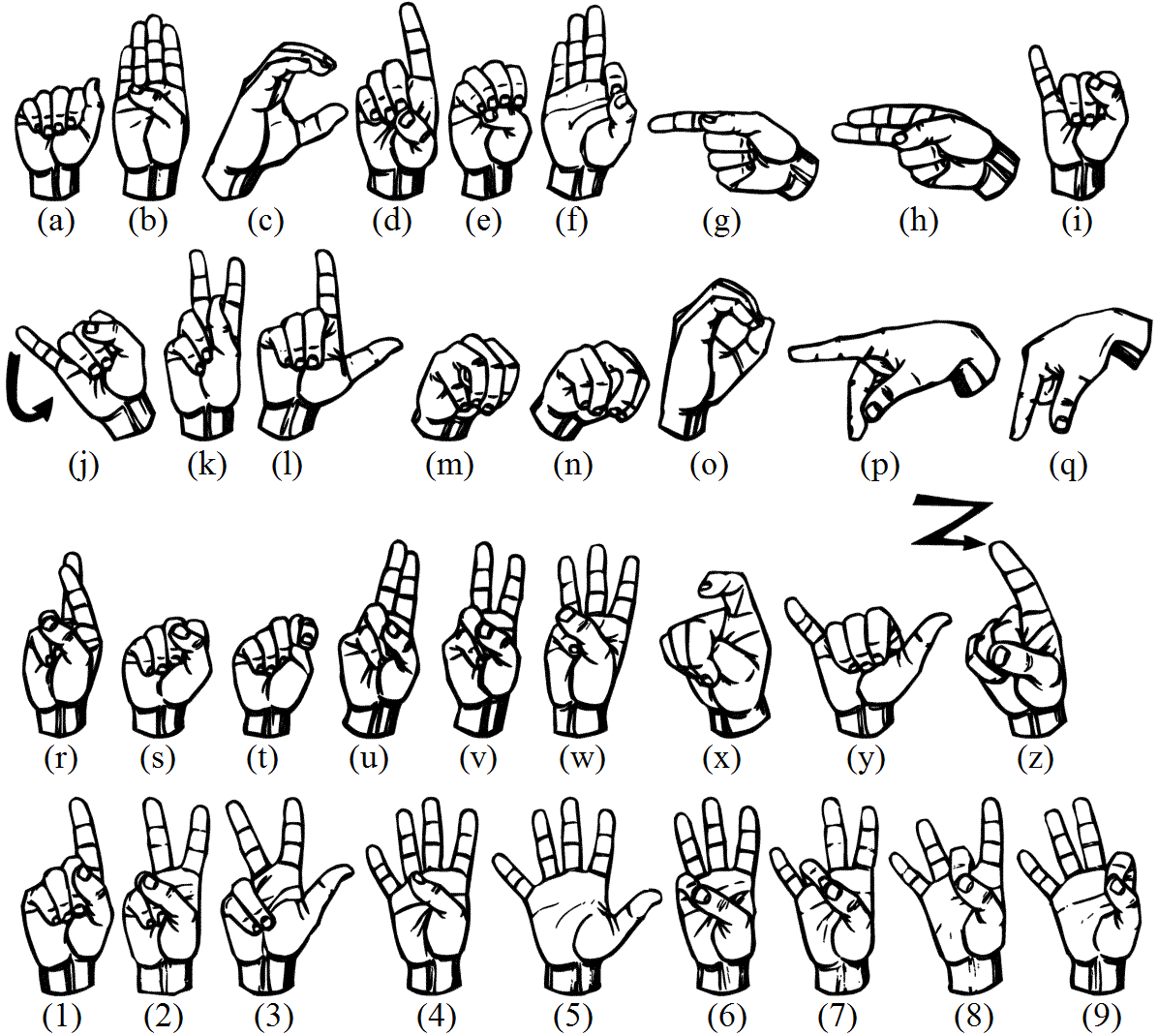}
\end{center}
   \caption{ASL fingerspelling alphabets and numbers~\cite{fing}. We follow the real demonstrations of formal signs on~\cite{wiki} to collect our dataset. The demonstration images are also available on our repository.}
\label{fig:fingerspelling}
\end{figure}

In this paper, we focus on static fingerspelling in American Sign Language (ASL) which is a small, but important part of sign language recognition. This is a small set of sign languages as shown in figure \ref{fig:fingerspelling}, but is used in many situations in conveying names, addresses, brands, and so on. Static fingerspelling is still challenging because of visually similar yet different signs. For example, some of the signs are only distinguished by the position of thumb. Also, a large variation occurs from different camera viewpoint and different signers.

Depth sensors enable us to capture additional information to improve accuracy and/or processing time. Also, with recent improvement of GPU, CNNs have been employed to many computer vision problems. Therefore, we take advantage of a depth sensor and convolutional neural networks to achieve a real-time and accurate sign language recognition system.

\section{Related Work}
Although gesture recognition only considers well specified hand gestures, some approaches are related to sign language recognition. Nagi \etal ~\cite{nagi11} proposed a gesture recognition system for human-robot interaction using CNNs. Van den Bergh \etal ~\cite{berg11} proposed a hand gesture recognition system using Haar wavelets and database searching. The system extracts features using Haar wavelets and classifies input image by finding the nearest match in the database. Although both systems show good results, these methods consider only six gesture classes. 

Different sign languages are used in different countries or regions. There have been efforts towards sign language recognition systems other than ASL as well.  
Pigou \etal ~\cite{pigo15} proposed an Italian sign language recognition system using CNNs. Although they reported 95.68\% accuracy for 20 classes, they mentioned that users in test set can be in training set and/or validation set. 
Liwicki \etal ~\cite{liwi09} described a British Sign Language recognition system that understands fingerspelled words from video. The system first recognized letters using Histogram of Gradients (HOG) descriptors. Then that recognized words using Hidden Markov Models (HMM). That system is different from recognizing a single fingerspelling. The dataset in use corresponded to a single signer. 

ASL sentence recognition and verification has also been explored. Zafrulla \etal ~\cite{zafr11} proposed a system which recognizes a sentence of three to five words. The word should be one of 19 signs in their dictionary. They also used Hidden Markov Models on extracted features. 

Our work belongs to the category of ASL fingerspelling recognition systems.
Isaac \etal~\cite{isaa04} proposed an ASL fingerspelling recognition system based on neural networks applied to wavelet features.
They reported a recognition rate of 99\%. However, they did not specify the size of the dataset and the number of different subjects.
Later, Pugeault \etal ~\cite{puge11} proposed a real-time ASL fingerspelling recognition system using Gabor filters and random forest. 
%The system consisted of an interactive user interface allowing the signer to select between ambiguous detection.
Their system recognizes 24 different ASL fingerspelling for alphabets. %Their data is collected from five subjects. 
They collected dataset from five subjects and reported a recognition rate of 75\% using both color and depth, 73\% using only color, and 69\% using only depth. 
Although Pugeault \etal ~\cite{puge11} reported that combination of color and depth improves the recognition rate, we only use depth to achieve better consistency to illumination changes and skin pigment differences and to avoid calibration process for general users.
Kuznetsova \etal ~\cite{kuzn13} also proposed a real-time ASL fingerspelling recognition system using multi-layered random forest. They reported and Dong \etal ~\cite{dong15} also analyzed that they achieved 87\% accuracy for the subjects whom the system has been trained on and 57\% accuracy for new subject.
Very recently, Dong \etal ~\cite{dong15} proposed an ASL alphabet recognition system. They first localized hand joint positions using random forest and hierarchical mode-seeking method. Then the system recognized ASL signs by applying random forest classifier to joint angle vector. They reported 90\% accuracy for the subjects whom the system has been trained on and 70\% accuracy for new signers.

This paper differs from previous works in several ways. First, to the best of our knowledge, ours is the first fingerspelling recognition system to classify total 31 alphabets and numbers compared with the state of the art approach to classify only 24 classes reported in the literature.
Second, we extract features by fine-tuning convolutional neural network parameters which are pre-trained for image classification task using 1.28 millions of color images~\cite{ilsvrc}.
Moreover, it achieves both real-time and the state of the art accuracy across different users. 
%Most of the previous works did not test their system for different signers whom the system has not been trained on.
Our contribution also includes providing publicly available dataset which is currently limited both in quantity and quality.
\section{Method}
\subsection{Dataset}
We have collected 31,000 depth maps using a depth sensor, Creative Senz3D camera of the resolution of 320 $\times$ 240. The dataset consists of 1,000 images for each of the 31 different hand signs from five subjects. 31 hand signs include all the fingerspellings of both alphabets and numbers except J and Z which require temporal information for classification. Since (2/V) and (6/W) are differentiated based on context, we have only one class to represent both one alphabet and one number. Although some informal signs are clearer and easier to recognize, we follow formal signs to avoid ambiguity between signers~\cite{wiki}. To collect dataset from various viewpoint, the dataset is collected while subjects are moving their hand around both on image plane and along z-axis.
\begin{figure}[t]
\begin{center}
   \includegraphics[width=0.55\linewidth]{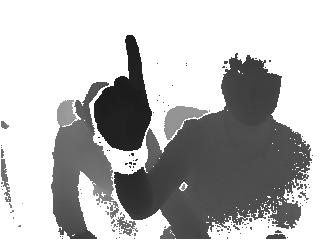}
\end{center}
   \caption{Captured image before pre-processing. The hand is convincingly the closest object according to the captured depth map and there is a clear depth void around the hand which can be exploited for hand segmentation using connected component from the area with the closest depth.}
\label{fig:original_input}
\end{figure}
\begin{figure}[t]
\begin{center}
   \includegraphics[width=1.0\linewidth]{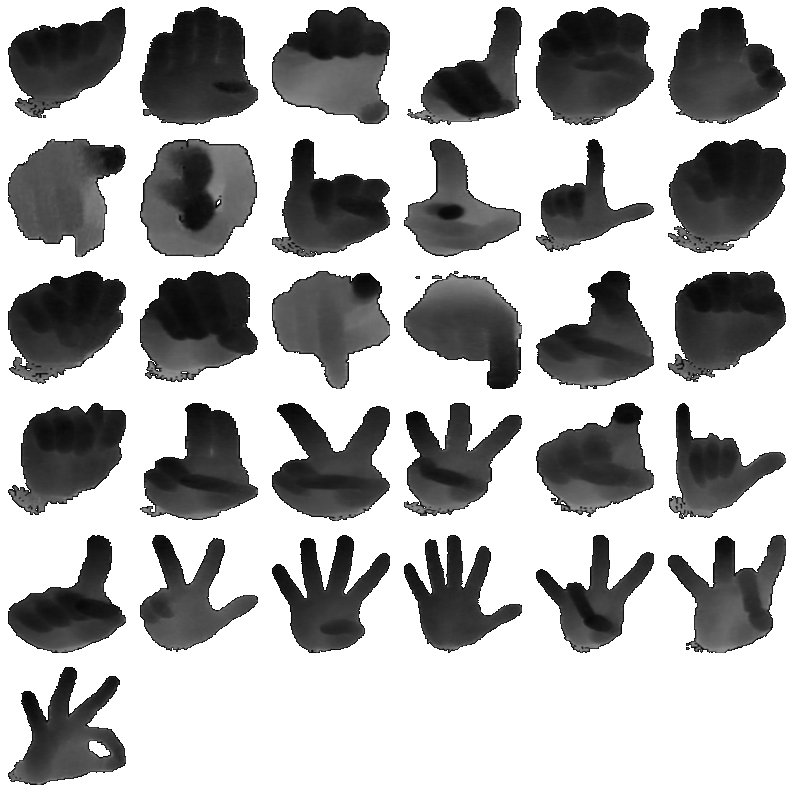}
\end{center}
   \caption{Examples of pre-processed dataset from A to Z and from 1 to 9. In real dataset, background is set to zero. }
\label{fig:dataset}
\end{figure}
\subsection{Hand Segmentation}
We assume that the closest object from camera is the user'’s hand. This assumption is valid in fingerspelling and most of gesture recognition tasks. In addition, we use a black wrist band to get depth voids around wrist, since depth sensor cannot capture depth from black objects well. Figure \ref{fig:original_input} shows one example of the captured depth image, where the hand is convincingly the closest object according to the captured depth map and there is a depth void around the hand. Hand segmentation thus ends up in finding the connected components from this closest region of the depth image. This strategy provides a very simple and effective real-time hand segmentation. Figure \ref{fig:dataset} shows segmented hand depth image samples for the 31 signs including alphabets and digits generated using this method. We find a bounding box of hand region and scale it to 227 $\times$ 227. Then we include redundant 14 or 15 pixels at each edge of the bounding box to make it the same input size of 256 $\times$ 256 described by Krizhevsky \etal ~\cite{kriz12} and thus take the differences of different hand segmentation results into account.  

\subsection{Classification}
\textbf{Architecture :}
We use Caffe~\cite{jia13} implementation (CaffeNet) of the CNNs which is almost equivalent to AlexNet ~\cite{kriz12}. The architecture consists of five convolution layers, three max-pooling layers, and three fully connected layers. After each convolution layer or fully connected layer except the last one, rectified linear unit layer is followed. For details, we will upload the architecture and also readers can refer to CaffeNet/Caffe~\cite{jia13} and AlexNet~\cite{kriz12}.

\textbf{Feature extraction :}
We extract a 4096-dimensional feature (final fully connected layer feature) vector from each pre-processed depth image using the aforementioned architecture. First, we subtract the mean image from each of the sample training/validation/test image. Then the mean-subtracted image is forward-propagated to extract features.

\textbf{Training :}
We train and test neural networks in five different operating modes. These five cases can be looked upon from different perspectives. One way to look at it is from the pre-training perspective and the second way is how we deal with the training/testing data separation for different subjects. In the former case, we categorize the operating modes into two categories, namely re-training and fine-tuning. For re-training, the model is re-trained from randomly generated weights using the collected fingerspelling data. In fine-tuning, we pre-train the CNNs using a large ILSVRC2012 classification dataset~\cite{ilsvrc}; then we fine-tune the network weights for fingerspelling classification with the same architecture except the last layer which is replaced by 31 output classes. From the subjects' data separation perspective, in one case, we do not separate the subjects in training, validation, and testing and in the second scenario, we use data from different subjects for training, validation, and testing.

\begin{figure}[t]
\begin{center}
   \includegraphics[width=1.0\linewidth]{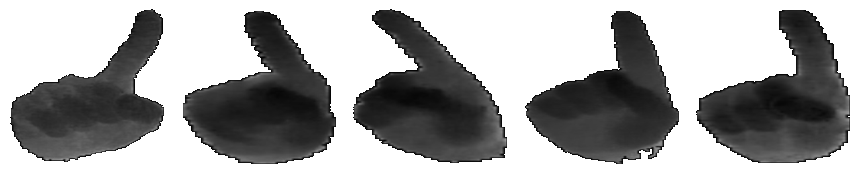}
\end{center}
   \caption{The collected data for the same meaning. It shows the importance of consideration of different signers and viewpoint variation.}
\label{fig:signer}
\end{figure}

\begin{table*}
\begin{center}
\begin{tabu} to \textwidth {X||c|c|c?c?c?c}
\Xhline{3\arrayrulewidth}
Method & Class type & \# of class & \# of subj. & Test w/ diff. & Input & Accur.(\%) \\
\hline\hline
Nagi \etal ~\cite{nagi11} & Gesture & 6 & - & No & Color & 96 \\
%Van den Bergh \etal ~\cite{berg11} & Gesture & 6 & - & No & Color & 99.54 \% \\
%Van den Bergh \etal ~\cite{berg11} & Gesture & 6 & - & No & Depth & 99.07 \% \\
Van den Bergh \etal ~\cite{berg11} & Gesture & 6 & - & No & Color \& Depth & 99.54 \\
Isaacs \etal ~\cite{isaa04} & Alphabets & 24 & - & - & Color &99.9 \\
Pugeault \etal ~\cite{puge11} & Alphabets & 24 & 5 & - &Color & 73  \\
Pugeault \etal ~\cite{puge11} & Alphabets & 24 & 5 & - &Depth & 69 \\
Pugeault \etal ~\cite{puge11} & Alphabets & 24 & 5 & - & Color \& Depth & 75 \\
Kuznetsova \etal ~\cite{kuzn13} (50/50)\% & Alphabets & 24 & 5 & No & Depth & 87 \\
Kuznetsova \etal ~\cite{kuzn13} (4/1)& Alphabets & 24 & 5 & {\cellcolor{lightgray}}Yes & Depth & {\cellcolor{lightgray}}57 \\
Dong \etal ~\cite{dong15} (50/50)\% & Alphabets & 24 & 5 & No & Depth & 90 \\
Dong \etal ~\cite{dong15} (4/1)& Alphabets & 24 & 5 & {\cellcolor{lightgray}}Yes & Depth & {\cellcolor{lightgray}}70 \\
Ours (re-training) (50/25/25)\% & Alph. \& Digit & 31 & 5 & No & Depth & 99.99 \\
Ours (re-training) (3/1/1) & Alph. \& Digit & 31 & 5 & {\cellcolor{lightgray}}Yes & Depth & {\cellcolor{lightgray}}75.18  \\
Ours (re-training) (4/1) & Alph. \& Digit & 31 & 5 & {\cellcolor{lightgray}}Yes & Depth & {\cellcolor{lightgray}}78.39  \\
Ours (fine-tuning) (3/1/1) & Alph. \& Digit & 31 & 5 & {\cellcolor{lightgray}}Yes & Depth & {\cellcolor{lightgray}}83.58  \\
Ours (fine-tuning) (4/1) & Alph. \& Digit & 31 & 5 & {\cellcolor{lightgray}}Yes & Depth & {\cellcolor{lightgray}}85.49  \\
\Xhline{4\arrayrulewidth}
\end{tabu}
\end{center}
\caption{Comparison. Gesture and ASL fingerspelling recognition systems are compared. Test with different subject means that the signer in test set is excluded from training/validation set. The corresponding results are highlighted in light gray. (a/b/c)\% represents the portion of dataset for (training/validation/test). (a/b/c) and (a/b) represent the number of subjects in (training/validation/test) and (training/test). }
\label{tab:comp}
\end{table*}

\begin{table*}
\begin{center}
\begin{tabu} to 1.0\textwidth {l|| X[c]?X[c]?X[c]?X[c]?X[c]?X[c]?X[c]?X[c]?X[c]?X[c]?X[c]}
\Xhline{3\arrayrulewidth}
  & A & B & C & D & E & F & G & H & I & K & L \\
Method & M & N & O & P & Q & R & S & T & U & V & W \\
 & X & Y & 1 & 3 & 4 & 5 & 7 & 8 & 9 & - & - \\ 
\hline\hline
Pugeault \etal~\cite{puge11} & 75 & 83 & 57 & {\cellcolor{lightgray}} 37 & 63 & \cellcolor{lightgray}35 & 60 & 80 & 73 & {\cellcolor{lightgray}}43 & 87 \\ 
Color \& Depth & {\cellcolor{lightgray}} 17 & {\cellcolor{lightgray}} 23 & {\cellcolor{lightgray}} 13 & 57 & 77 
 & 63 & {\cellcolor{lightgray}}17 & {\cellcolor{lightgray}} 7 & 67 & 87 & 53 \\
 & \cellcolor{lightgray}20 & 77 & - & - & - & - & - & - & - & - & -\\
\hline

Ours & 82.7 & \textbf{94.9} & 83.3 & 85.9 & \cellcolor{lightgray}{45.7} & 86.6 & 86.1 & 81.8 & 72.5 & 86.7 & \textbf{93.4} \\
(re-training) & \cellcolor{lightgray}{42.1} & \cellcolor{lightgray}{32.6} & 73.1 & 85.4 & 70.8 
 & 58.9 & 73.6 & \cellcolor{lightgray}{10.1} & 61.7 & \textbf{93.0} & 80.2 \\
 (3/1/1) & 66.9 & 51.0 & \textbf{98.6} & 80.3 & \textbf{92.7} & \textbf{95.7} & \textbf{92.7} & 79.6 & \textbf{92.6} & - & -\\
\hline

Ours & 85.8 & \textbf{95.4} & \textbf{91.7} & \textbf{91.4} & \cellcolor{lightgray}{43.0} & 83.6 & 79.9 & 81.5 & 78.7 & \textbf{94.4} & \textbf{98.4} \\
(re-training) & \cellcolor{lightgray}{46.8} & \cellcolor{lightgray}{28.4} & 74.9 & 79.0 & 70.2 & 69.4 & 73.2 & \cellcolor{lightgray}{4.8} & 70.5 & \textbf{98.1} & 87.6 \\
(4/1)& 75.8 & 86.0 & \textbf{97.7} & 79.5 & \textbf{92.2} & \textbf{96.3} & \textbf{98.1} & 80.2 & \textbf{97.6} & - & -\\
\hline

Ours & 84.2 & \textbf{94.3} & 86.9 & 89.7 & 87.1 & \textbf{92.3} & 88.0 & 85.2 & \textbf{90.5} & 83.0 & \textbf{99.7} \\
(fine-tuning) & 62.2 & 59.0 & 69.4 & 82.9 & 82.6 & 80.9 & 57.6 & 55.3 & \textbf{92.2} & \textbf{94.5} & 83.7 \\
(3/1/1)& 72.0 & 76.9 & \textbf{99.0} & 77.9 & \textbf{91.0} & \textbf{98.1} & \textbf{95.4} & 82.0 & \textbf{98.1} & - & -\\
\hline

Ours & 89.5 & \textbf{97.1} & \textbf{93.2} & 88.4 & 85.7 & \textbf{93.8} & 84.9 & 86.1 & \textbf{95.6} & \textbf{90.0} & \textbf{99.4} \\
(fine-tuning) & 59.9 & 65.1 & 69.2 & 80.2 & 85.7 & 85.2 & 53.7 & 60.6 & \textbf{96.1} & \textbf{98.2} & 87.0 \\
(4/1) &75.3 & 81.9 & \textbf{99.0} & 82.1 & \textbf{92.1} & \textbf{97.5} & \textbf{95.9} & 82.5 & \textbf{99.3} & - & -\\
\Xhline{4\arrayrulewidth}
\end{tabu}
\end{center}
\caption{Detailed results. Accuracy less than 50\% is highlighted in light gray. Bold entries correspond to accuracy higher than 90\%.}
\label{tab:detail}
\end{table*}

\section{Experimental Results}

As mentioned in Sec. 3.3, we train and test for five experimental settings. The results are compared with other systems in table~\ref{tab:comp}. Our system achieves 99.99\% accuracy when training and validation data have samples corresponding to the test subject. In this experiment, 50\%, 25\%, and 25\% of dataset is used for training, validation, and testing. We achieve 75.18\% and 78.39\% for regular training (re-training) and 83.58\% and 85.49\% for fine-tuning. This shows that fine-tuning outperforms re-training about 7$\sim$8\% in this depth image dataset even though the nature of the pre-training dataset (ILSVRC2012) is different. For each case, the former represents the average result of training with three subjects, validation with one subject, and test with one subject. The latter represents the average result of training with four subjects and test with one subject. We considered all possible combinations of subjects for training, validation, and test and the final reported accuracy is the average of all. For the latter case, although we use the same training parameters (e.g. the number of iterations) for all combinations, the performance is improved about 2$\sim$3\%. It shows that by increasing the number of subjects in training, the system's performance for new subject has high possibility of improvement. The number of iterations in the training phase is fixed to 8000 and 4000 for re-training and fine-tuning respectively. Overall, our system achieves about 10$\sim$15\% improvement compare to previous state of the art result even with more number of classes and subjects. Moreover, the processing time is about 3 ms for the prediction of a single image using Nvidia GeForce GTX Titan.
% * <subarna.tripathi@gmail.com> 2015-07-07T18:02:39.835Z:
%
%  add a little description for the failure cases and refer to the correct figure. 
%
% : replace figure by explaining which cases have low accuracy
%
Table~\ref{tab:detail} shows accuracy for each alphabet or number. It shows that even with more number of classes, the accuracy of each class is higher than or similar to previous state of the art result. Therefore, it is obvious we will achieve better result with the same number of classes of comparing with the other works. On the other hand, the table shows that (E,M,N,T) has low accuracy for both our method and others since the letters are only differentiated by thumb position.

Previous gesture or ASL fingerspelling recognition systems considered only six gestures or 24 signs respectively. We increase the number of signs to 31 to accommodate most of the signs. Also, some previous methods did not separate subjects in training, validation, and testing. Experimenting with training and testing dataset from different subjects is important because then only the system's performance for a new subject can be measured. Therefore, we include experiments and report results where training and testing subjects are separated to demonstrate how the trained model can be used for anonymous subject.
% * <subarna.tripathi@gmail.com> 2015-07-07T19:24:31.626Z:
%
%  why not RGBD ? -- do we need to mention something on that ?
% 
%  -- : I think it will explain (from Section 2)
%
Lastly, although Pugeault \etal ~\cite{puge11} reported that combination of color and depth improves the recognition rate, we only use depth to achieve better consistency to illumination changes and skin pigment differences and to avoid calibration process for general users.
%------------------------------------------------------------------------
\section{Conclusion}
We show the efficacy of using convolutional neural networks and a depth sensor for ASL fingerspelling recognition system. We collect and share the dataset of depth images for ASL fingerspelling system. Our approach of classifying 31 signs of alphabets and numbers using depth image and CNNs achieves real-time performance and state-of-the-art accuracy even for different signers. We conclude that pre-training from auxiliary task of image classification from color images is helpful for apparently different type of input data such as depth image. 
%In future, we will extend the system to recognize alphabets or words that require temporal information. 
The trained model and dataset is available on our repository {\url{https://github.com/byeongkeun-kang/FingerspellingRecognition}}.

{\small
\bibliographystyle{ieee}
\bibliography{egpaper_for_camera_ready}
}

\end{document}